\documentclass[journal]{IEEEtran}
\usepackage{cite}

  \usepackage[pdftex]{graphicx}

\usepackage{amsmath}
\usepackage{algorithmic}

\usepackage{array}

\usepackage[caption=false,font=footnotesize]{subfig}
\usepackage{url}

\usepackage{bm}
\usepackage{amssymb}
\usepackage{algorithm}

\newboolean{include-notes}
\setboolean{include-notes}{false}
\newcommand{\todo}[1]{\ifthenelse{\boolean{include-notes}}%
 {\textcolor{red}{\textbf{TODO: #1}}}{}}
\newcommand{\note}[1]{\ifthenelse{\boolean{include-notes}}%
 {\textcolor{blue}{\textbf{#1}}}{}}

\newcommand{\figref}[1]{Fig. \ref{#1}} 
\newcommand{\algref}[1]{Algorithm \ref{#1}} 
\newcommand{\eqnref}[1]{Eqn. \ref{#1}} 

\DeclareFontFamily{OT1}{pzc}{}
\DeclareFontShape{OT1}{pzc}{m}{it}{<-> s * [1.10] pzcmi7t}{}
\DeclareMathAlphabet{\mathpzc}{OT1}{pzc}{m}{it}

\begin{document}
%
%
%
%

\title{LaPlaSS: Latent Space Planning for \\ Stochastic Systems}
\author{Marlyse~Reeves, Brian~Williams
    \thanks{M. Reeves and B. Williams are with the Massachusetts Institute of Technology. Contact: \{mreeves, williams\}@mit.edu}}%

\maketitle

\begin{abstract}

Autonomous mobile agents often operate in hazardous environments, necessitating an awareness of safety. These agents can have non-linear, stochastic dynamics that must be considered during planning to guarantee bounded risk. Most state of the art methods require closed-form dynamics to verify plan correctness and safety however modern robotic systems often have dynamics that are learned from data. Thus, there is a need to perform efficient trajectory planning with guarantees on risk for agents without known dynamics models. We propose a "generate-and-test" approach to risk-bounded planning in which a planner generates a candidate trajectory using an approximate linear dynamics model and a validator assesses the risk of the trajectory, computing additional safety constraints for the planner if the candidate does not satisfy the desired risk bound. To acquire the approximate model, we use a variational autoencoder to learn a latent linear dynamics model and encode the planning problem into the latent space to generate the candidate trajectory. The VAE also serves to sample trajectories around the candidate to use in the validator. We demonstrate that our algorithm, LaPlaSS, is able to generate trajectory plans with bounded risk for a real-world agent with learned dynamics and is an order of magnitude more efficient than the state of the art. 
\end{abstract}


%
\IEEEpeerreviewmaketitle

\section{Introduction}

Autonomous, stochastic mobile agents are being tasked to perform increasingly complex missions that require agents to achieve multiple goals over time in hazardous environments. As other work has argued \cite{Li2008, Chen2021, Garrett2021, Scotty2018, Jin2021}, it is important to consider the coupling between discrete tasks and continuous motions when planning for these agents by integrating activity and trajectory planning. However, existing approaches for combined activity and trajectory planning assume access to stochastic agent dynamics as a closed-form expression to provide guarantees on the correctness and safety of the plan. In reality, state of the art robotic systems often have uncertain dynamics that are arbitrarily complex and hence learned from data. In this work, we focus on addressing the need for risk-bounded trajectory planning for stochastic agents with learned dynamics models. Additionally, to ensure adaptability, our approach must be efficient enough allow for online trajectory planning.

The state of the art in risk-bounded trajectory planning for stochastic nonlinear agents formulates the planning problem as into a standard nonlinear optimization problem by converting chance constraints and stochastic dynamics constraints into deterministic constraints on moments \cite{weiqiao2022}. While this approach provides precise trajectories and tight bounds on risk, it is too slow for online trajectory planning. Other approaches successfully leverage efficient convex optimization techniques by reframing the stochastic planning problem as a deterministic planning problem using a safety margin \cite{ono2008b, blackmore2009}. While these approaches are much more efficient, they assume linear dynamics models, an abstraction that can lead to under-conservative trajectories for a nonlinear agent. To address these challenges, our approach must balance efficiency and accuracy.

The methods discussed so far operate on closed-form dynamics models which might not be known. Advances in deep learning techniques allow accurate dynamics models to be learned from data. Autoencoders and their variants are used to learn dynamics models of high-dimensional, non-linear systems \cite{jonschkowski2014, agostini2020, lazzara2022}. While these methods are powerful for inference tasks like prediction, they lack a symbolic representation of the dynamics (e.g. closed-form dynamics equations) suitable for trajectory optimization. To address this limitation, other approaches learn linear embeddings of the dynamics by constraining the latent space to exhibit properties of common state space models such as Bayes filters, Kalman filters, iLQG control formulations, and Koopman operators \cite{watter2015, fraccaro2017, karl2017, lusch2018, li2020}. We draw inspiration from this body of work to learn a linear latent dynamics model from trajectory data for convex trajectory optimization.

Finally, many trajectory planning problems require the satisfaction of user-defined state and control constraints in addition to respecting the agent dynamics. Prior work aims to learn latent dynamics models from high dimensional inputs and perform planning in the latent space \cite{hafner2018,  nguyen2021, olesen2021}. However, these planning approaches are reward-based and do not provide guarantees on constraint satisfaction. Other research has focused on exploring a lower dimensional latent space to find optimal solutions to constrained optimization problems but these explorations are only constrained by boundary conditions (i.e. operating extremes of the system) as opposed to arbitrary constraints \cite{engel2018, lew2021, park2022}. \cite{morton2019} takes an approach most similar to ours, using a variational autoencoder to obtain a distribution over possible outcomes given uncertain dynamics. While their control approach models stochasticity, it does not attempt to bound the risk incurred over the planned trajectory.

We introduce Latent Space Planning for Stochastic Systems (LaPlaSS), an algorithm for risk-bounded planning for non-linear stochastic agents with learned dynamics. We employ a "generate-and-test" approach to bounding risk in which a planner generates a candidate trajectory using an approximate linear dynamics model and a validator assesses the risk of the trajectory. The validator samples trajectories around the candidate control trajectory, generating a \textit{probabilistic flow tube}. Leveraging fast risk-assessment techniques, the validator evaluates the risk of the candidate and computes new safety constraints to pass to the trajectory planner to refine the candidate trajectory. To exploit powerful convex optimization techniques for trajectory planning, we train a variational autoencoder approach to learn an accurate dynamics model and a linear embedded latent space. We encode the planning problem into latent space, utilizing the linear latent dynamics to perform trajectory optimization, and decode the result to obtain a candidate trajectory in real space. We use an approximation of the state and control constraints to ensure convexity of the constraints in the latent space. Our method iterates between trajectory planning and trajectory validation until a \textit{safe} trajectory (i.e. a trajectory that satisfies the provided risk bound) is found or the problem is deemed infeasible.
\section{Problem Statement}

Consider an autonomous mobile agent with state space $\mathcal{X} \subseteq \mathbb{R}^n$ and control space $\mathcal{U} \subseteq \mathbb{R}^m$, where $n,m \in \mathbb{N}^+$. The system dynamics is given by
\begin{equation}
    \bm{x}_{t+1} = f(\bm{x}_t, \bm{u}_t, \omega_t, \Delta t)
    \label{eqn:dyn}
\end{equation}
where $\bm{x} \in \mathcal{X}$, $\bm{u} \in \mathcal{U}$ are the state and control input at the current time step, $\omega$ is the noise, and $\Delta t$ is the fixed time step. The initial state $\bm{x}_0$ is given by a known probability distribution, $I(\bm{x}_0)$. The goal region, $\bm{P}_{goal}$, is a convex polytope in $\mathcal{X}$.  The obstacles in the environment, denoted $\mathcal{O}_k, k = 1,\cdots,K$, where $K \in \mathbb{N}^+$, are ellipsoidal, parameterized by a constant, positive semi-definite matrix $Q_k$ such that \todo{add footnote?}
\begin{equation}
    \{\mathbf{x}: Q_(\mathbf{x}) \leq 1\}, \, Q_k \in \mathcal{S}^2_{++}
\end{equation}
We define \textit{risk} as the probability of collision with any obstacle at any time step. The goal of risk-bounded trajectory planning is to find a control trajectory $u_0,...,u_T$ that minimizes a cost function $J(\bm{x}, \bm{u})$ and bounds the risk below the provided bound, $\delta$. More formally, the risk-bounded trajectory planning problem can be formulated as an optimization problem 

\textbf{Problem 1. } Risk-bounded Trajectory Optimization
\begin{align*}
    \underset{\bm{u}}{\text{min }} & J(\bm{x}, \bm{u}) \\
    \text{s.t. } &\bm{x}(0) \sim I(\bm{x}_0)\\
    & \bm{x}_{t+1} = f(\bm{x}_t, \bm{u}_t, \omega_t, \Delta t) \\
    & H\bm{x} + L\bm{u} \leq \bm{g} \\
    &\bm{x}(T) \in \bm{P}_{goal} \\
    & \text{Prob}(\bm{x} \in \mathcal{O}) \leq \delta
\end{align*}

where $H\bm{x} + L\bm{u} \leq \bm{g}$ represents a set of arbitrary, convex state constraints.

Recall, however, that the autonomous mobile agents that we are addressing in this work do not have an explicit dynamics function like \eqref{eqn:dyn}. Instead, we have state and control trajectory data of the agent from which we must learn a dynamics model, $f'$. Thus, our problem can be formulated as

\textbf{Problem 2. } Risk-bounded Trajectory Optimization with Learned Dynamics

Given time series data as $N$ state and control trajectories of length $M$, 
$$ \bm{X} = \begin{matrix}
\boldsymbol{x}_{0,0} & \cdots & \boldsymbol{x}_{0,M} \\ \boldsymbol{x}_{1,0} & \cdots & \boldsymbol{x}_{1,M} \\
\vdots & \ddots & \vdots \\
\boldsymbol{x}_{N,1} & \cdots & \boldsymbol{x}_{N,M}
\end{matrix}, \, \bm{U} = 
\begin{matrix}
\boldsymbol{u}_{0,0} & \cdots & \boldsymbol{u}_{0,M} \\
\boldsymbol{u}_{1,0} & \cdots & \boldsymbol{u}_{1,M} \\
\vdots & \ddots & \vdots \\
\boldsymbol{u}_{N,1} & \cdots & \boldsymbol{u}_{N,M}
\end{matrix}$$
learn a dynamics function $f'$ then solve Problem 1.

In this paper, we assume that the time series data adequately covers the operating regime of the autonomous agent.

\section{Approach Overview}
\begin{figure}
    \centering
    \includegraphics[width=0.45\textwidth]{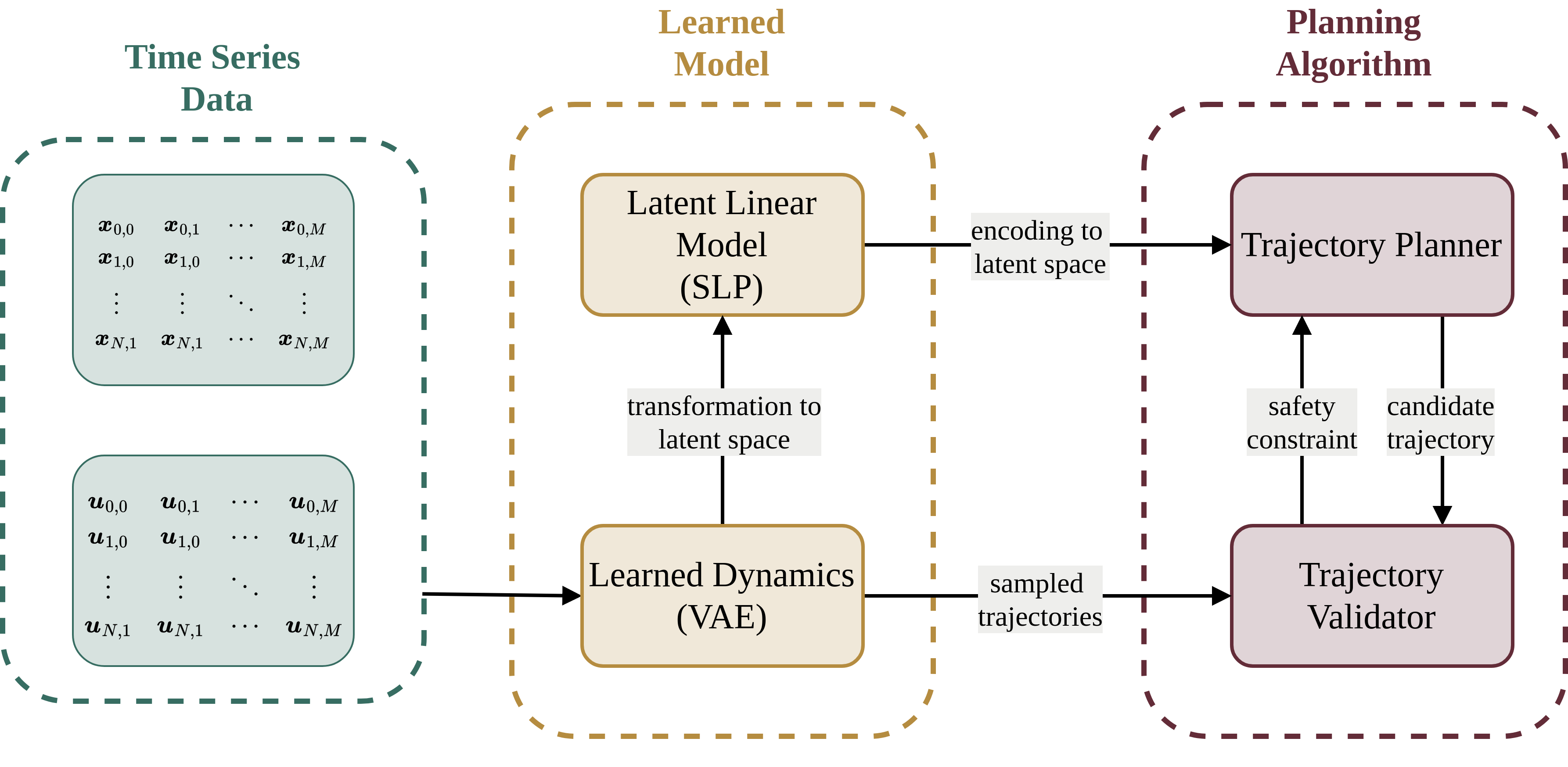}
    \caption{High-level Diagram of Approach}
    \label{fig:approach-overview}
\end{figure}

We propose a "generate-and-test" approach to risk-bounded planning in which a planner generates a candidate trajectory using an approximate linear dynamics model and a validator assesses the risk of the trajectory, computing additional safety constraints for the planner if the candidate does not satisfy the desired risk bound. To acquire the approximate model, we use a variational autoencoder to learn a latent linear dynamics model and encode the planning problem into the latent space to generate the candidate trajectory. An overview of our approach is visualized in \figref{fig:approach-overview}. We begin by discussing the iterative risk-bounded planning algorithm at a high-level then outline how we learn a model for planning from trajectory data.

\subsubsection{Planning Algorithm}
First, the trajectory planner generates a candidate control trajectory from the initial state to the goal region using a relaxed trajectory optimization formulation. The trajectory planner uses a linear approximation of the agent dynamics and assumes the agent is deterministic but does consider the obstacles and state and control constraints. The trajectory planner passes the candidate trajectory to the trajectory validator.

The trajectory validator samples trajectories around the candidate using an accurate, stochastic dynamics model. The validator fits a probability distribution to the sampled states at each time step. We refer to this sequence of distributions as a \textit{probabilistic flow tube} (PFT). The PFT is a representation of the possible states the agent could enter as it executes the candidate trajectory. Leveraging the fast risk-assessment technique from \cite{wang2020}, we evaluate the probability of collision with the obstacles in the environment using the PFT. If the risk of collision is below the risk bound $\delta$, we return the candidate trajectory as the solution. Otherwise, we compute a new convex constraint at the point in the trajectory with the highest collision probability. This safety constraint constrains the trajectory to avoid that high-risk state. The safety constraint is passed back to the trajectory planner to be including the trajectory optimization problem in the next iteration. Our planning algorithm terminates when a \textit{safe} (i.e. $\text{Prob}(\bm{x} \in \mathcal{O}) \leq \delta$) trajectory is found or the problem is deemed infeasible.

\subsubsection{Model Learning}
Because the autonomous agents we are considering do not have known closed-form dynamics, we must learn both an accurate stochastic dynamics model to generate trajectory samples for the trajectory validator and an approximate deterministic linear dynamics model for the trajectory planner. For the former, we use a variational autoencoder (VAE) model that captures the uncertainty in the dynamics by learning a parameterized latent distribution that fits the data \cite{Kingma2013}. Our approach uses two VAEs, one for the state and one for the control, that encode the state and control input vectors into latent distributions that are sampled to form the a latent state vector $\bm{z}_x$ and latent control vector $\bm{z}_u$, respectively. Given a current state $\bm{x}_t$ and control input $\bm{u}_t$, this learned dynamics model predicts the next state $\bm{x}_{t+1}$. 

In the latent space learned by the VAEs, we train a single layer perceptron (SLP) as a linear mapping, $\mathcal{K}$, from $\bm{z}_{x,t}$ and $\bm{z}_{u,t}$ to $\bm{z}_{x,t+1}$ such that
\begin{equation}
    \bm{z}_{x,t+1} = \mathcal{K} \cdot  \begin{bmatrix}
           \bm{z}_{x,t} \\
           \bm{z}_{u,t}
         \end{bmatrix}
\end{equation}

To use our learned linear dynamics in the trajectory planner, we must encode the trajectory optimization problem into the latent space. We use the VAEs to encode the initial state and goal region as well as the state and control constraints into the latent space. 
We then formulate a trajectory optimization problem in the latent space from the latent start state $\bm{z}_{x,0}$ to the latent goal region $\bm{z}_{P, goal}$ that respects the latent state and control constraints. Once the trajectory planner has found the optimal latent control trajectory $\bm{z}_u^*$, we decode the trajectory back into the real space. This decoded trajectory $\bm{u}^*$ is the candidate control trajectory passed to the trajectory validator.

In the following sections, we will describe the trajectory validator and learned model architecture in more detail.
\section{Trajectory Validation}
Given a candidate control trajectory $\bm{u}^*$, a set of obstacles $\mathcal{O}$, risk bound $\delta$, and an accurate dynamics model $f'$, the problem for the trajectory validator is to determine if $\text{Prob}(\bm{x} \in \mathcal{O}) \leq \delta$. If not, the trajectory validator must return a safety constraint, $l_{safety}$ to pass back to the trajectory planner. The algorithm for the trajectory validator is given in \algref{alg:validator}. 
\begin{algorithm}[tb]
\caption{Trajectory Validation}
\label{alg:validator}
\textbf{Input}: $\bm{u}^*$, a set of obstacles $\mathcal{O}$, risk bound $\delta$, and an accurate dynamics model $f'$\\
\textbf{Output}: safety constraint, $l_{safety}$ or \texttt{None} \\
\begin{algorithmic}[1] 
\STATE Let $risk=0.0$
\STATE $tube$ = \texttt{COMPUTE\_PFT}($f'$, $\bm{u}^*$)
\FOR{$o \in \mathcal{O}$}
\STATE $risk \mathrel{{+}{=}} $ \texttt{ASSESS\_RISK}($tube$, $o$)
\ENDFOR
\IF {$risk \leq \delta$}
\STATE \textbf{return} \texttt{None}
\ELSE
\STATE $o^* = \underset{o}{\text{argmax }} \texttt{ASSESS\_RISK}(tube, o)$
\STATE \textbf{return} \texttt{COMPUTE\_SAFETY\_CONSTRAINT}($o^*$)
\ENDIF
\end{algorithmic}
\end{algorithm}

\subsubsection{Generating a PFT}
\begin{figure*}
    \centering
    \includegraphics[width=\textwidth]{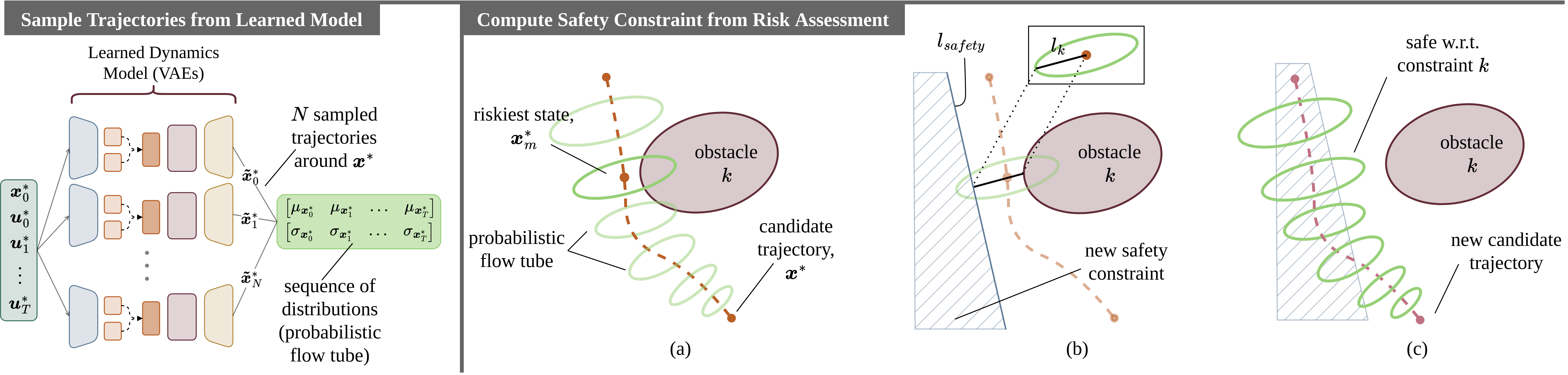}
    \caption{(left) Trajectories sampled from learned dynamics model to fit a sequence of distributions that form the probabilistic flow tube. (right) Illustration of safety constraint computation. (a) The "riskiest" state in the candidate trajectory after risk assessment using the PFT. (b) Computing the safety constraint using the confidence ellipse of the distribution at the riskiest state. (c) New candidate trajectory constrained by the safety constraint computed in the previous iteration.}
    \label{fig:validator}
\end{figure*}

First, the trajectory validator computes a probabilistic flow tube around the candidate trajectory using an accurate dynamics model. In our approach, the dynamics model $f'$ is the learned VAE model. Given an initial state $\bm{x}_0$ and a candidate control trajectory $\bm{u}^*$, the VAE model can predict the trajectory that results from executing $\bm{u}^*$. Because the VAE model captures the uncertainty in the agent dynamics via latent distributions that are sampled to obtain the latent vectors $\bm{z}$, applying the VAE to the candidate trajectory is equivalent to generating a trajectory sample around the candidate. The trajectory validator collects $N \sim 10^4$ samples then fits a Gaussian distribution for each time step. The result is a sequence of parameterized distributions  $[\begin{smallmatrix}
\mathcal{N}(\mu_0, \sigma_0), & \mathcal{N}(\mu_1, \sigma_1), & \cdots, & \mathcal{N}(\mu_T, \sigma_T)
\end{smallmatrix}]$ that is the probabilistic flow tube. \figref{fig:validator} (left) illustrates this trajectory sampling procedure.

\subsubsection{Fast Risk Assessment} Next, the trajectory validator computes the risk of collision using the probabilistic flow tube.

Let $\mathbf{x}_t \sim \mathcal{N}(\mu_t, \sigma_t)$ be a random vector, with an underlying Gaussian distribution given by the probabilistic flow tube representing the uncertain state of the system at time $t$. The risk, or probability of collision, associated with ellipsoidal obstacle $k$ across the whole time horizon $T$ is 
\begin{equation}
    \mathcal{R}_k :=  \mathbb{P} \left( \bigcup^T_{t=1} \{Q_k(\mathbf{x}_t) \leq 1\} \right)
    \label{prob_1}
\end{equation}
By the inclusion-exclusion principle, the probability in \eqnref{prob_1} can be computed as the sum of the probabilities of the marginal events (i.e. agent collides with obstacle $k$ at time $t$) and the probabilities of all possible intersections of events (i.e. agent collides with obstacle $k$ at time $t$ \& $t+1$, etc.)
\begin{equation}
    \mathcal{R} = \sum_{J \in \mathcal{P}([T])} (-1)^{|J| + 1} \mathbb{P}\left(\bigcap_{j \in J} \{Q_k(\mathbf{x}_t)\} \leq 1\right)
\end{equation}
where $P([T])$ is the power set of the set of time points in the horizon. We assume that the risk of collision is independent across time. In the time-independent case, the risk computation reduces to
\begin{equation}
    \mathcal{R} = 1 - \prod_{t \in [T]} \left(1 - \mathbb{P}\left(Q_k(\mathbf{x}_t) \leq 1\right) \right)
\end{equation}
Thus, the problem of risk assessment can be solved by computing the marginal probabilities of violation $\mathbb{P}\left(Q_k(\mathbf{x}_t) \leq 1\right)$ for each time $t$. This is exactly the CDF of $Q_k(\mathbf{x}_t$, which is a quadratic form in a multivariate Gaussian. While no closed form solutions to exactly evaluate these CDFs, fast approximation methods with bounded error exist, trading between accuracy and evaluation speed. In this work, we use the method of Liu-Tang-Zhang to get the probability of collision with each constraint at each time step in the PFT \cite{liu2009}.

\subsection{Computing Safety Constraints}

Suppose that the probabilistic flow tube sampled around the candidate trajectory has the largest probability of collision with obstacle $k$ at time $m$. This is the 'riskiest' portion of the trajectory (\figref{fig:validator} (right)(a)). Let $\mu_m$ and $\sigma_m$ be the parameters of the distribution around the candidate trajectory at time $m$. The eigenvectors of the covariance matrix $\Sigma_m$ correspond to direction of highest variance of the agent state. These eigenvectors form the axes of the \textit{confidence ellipsoid} (\figref{fig:validator} (right)(b)).

Next, the validator computes the point $\bm{x}_k$ on the boundary of obstacle $k$ closest to $\bm{x}^*_m$ in the direction of the largest axis of the confidence ellipsoid. With high probability, this is the direction in which more of the risk is being incurred. We create a hyperplane $l_{k}$ from $\bm{x}_k$ away from the obstacle (and towards the candidate trajectory) of length $\sqrt{\beta\lambda_1}$, where $\lambda_1$ is the largest eigenvalue of $\Sigma_m$ and $\beta$ is a tunable parameter (\figref{fig:validator} (right)(b)). We initialize $\beta = 4.6$, which corresponds to an 90\% confidence ellipse.

Finally, the validator computes a hyperplane $l_{safety}$ that is perpendicular to hyperplane $l_{k}$. We create a polytopal safety constraint from the intersection of $l_{safety}$ and the bounds of the environment and pass the constraint back to the trajectory planner. The trajectory planner constrains the trajectory to pass through the new safety constraint around time $t = m$ in the next iteration of the algorithm (\figref{fig:validator} (right)(c)). If the new candidate trajectory is safe with respect to risk bound $\delta$, we return the trajectory as the solution. Otherwise, we increase $\beta$ (which will push safety constraints farther away from the risk obstacle in the next iteration), generate a new safety constraint, and continue iterating on the candidate trajectory until a safe solution is found (or the problem is deemed infeasible).

\section{Trajectory Planning with Learned Models}
In this section, we detail how we learn, and subsequently plan with, dynamics models from trajectory data. Recall that we are given as input time series data as $N$ state and control trajectories of length $M$ that adequately cover the autonomous agent's operating regime. 

\begin{figure*}
    \centering
    \includegraphics[width=\textwidth]{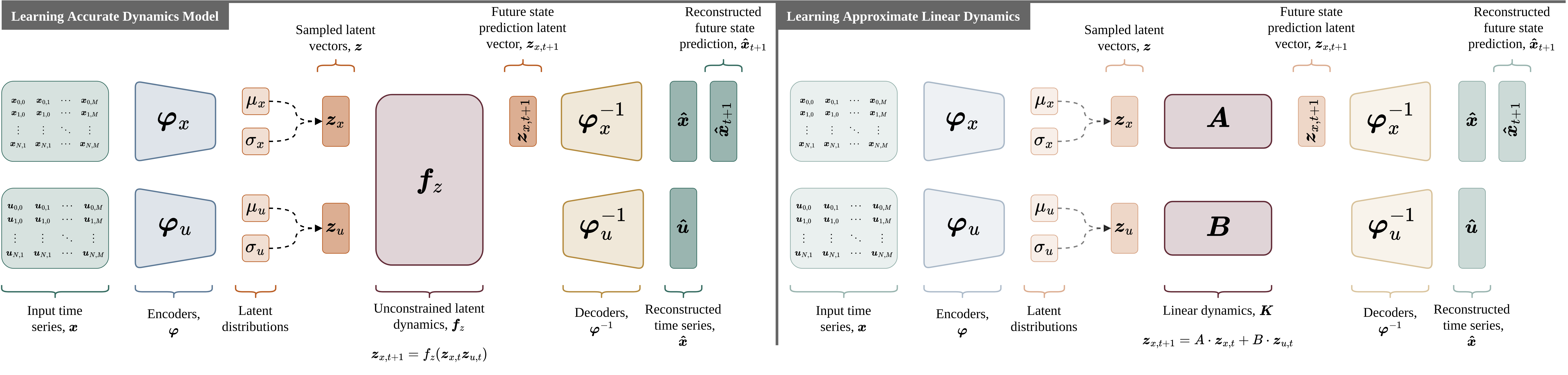}
    \caption{(left) Architecture for accurate learned dynamics. (right) Architecture for approximate linear latent dynamics.}
    \label{fig:learning}
\end{figure*}

\subsubsection{Learning Accurate Dynamics}
First, we learn a neural model that, given a current state $\bm{x}_t$ and control input $\bm{u}_t$, can predict the next state $\bm{x}_{t+1}$. Our network architecture consists of two deep variational autoencoders $\bm{\varphi}_x$ and $\bm{\varphi}_u$ for the state and control vectors respectively (\figref{fig:learning} (left)). These learn parameterized Gaussian latent distributions are sampled to obtain the latent state vector $\bm{z}_x \sim \mathcal{N}(\mu_x, \sigma_x)$ and latent control vector $\bm{z}_u \sim \mathcal{N}(\mu_u, \sigma_u)$. 

We train both VAEs jointly using a loss function given by
\begin{align}
    L_{VAE} = &\overbrace{\parallel \bm{x} - \bm{\hat{x}} \parallel}^{\text{reconstruction loss}}  + \,
    \overbrace{\parallel \bm{x}_{t+m} - \bm{\hat{x}}_{t+m} \parallel}^{\text{prediction loss}} \\ &+ \underbrace{D_{KL}(\mathcal{N}_x) + D_{KL}(\mathcal{N}_u)}_{\text{KL divergence}} \nonumber
\end{align}
where $m$ is the number of timesteps over which to enforce the prediction loss and is a tuned parameter. Note that each component of the loss function is weighted and weights are determined via tuning. We will use this VAE model to sample trajectories around a candidate trajectory in the trajectory validator.

\subsubsection{Learning Approximate Linear Dynamics}

Once we have learned an accurate dynamics model with an encoding from the real-space to the latent space, we learn an approximate linear dynamics model as a single layer perceptron in the latent space. The linear latent dynamics function $\mathcal{K}$ can be decomposed into to two linear functions $A$ and $B$ such that
\begin{equation}
    \bm{z}_{t+1} = \mathcal{K} \cdot \bm{z}_t = A \cdot \bm{z}_{x,t} + B \cdot \bm{z}_{u,t}
\end{equation}

The loss function operates only on the latent variables 
\begin{equation}
    L_{\mathcal{K}} = \overbrace{\parallel \bm{z}_{t+m} - K^m\bm{z}_t \parallel}^{\text{linearity loss}}
\end{equation}
Again, $m$ is the number of timesteps over which to enforce the linearity loss and is a tuned parameter. A diagram of this network architecture is shown in \figref{fig:learning} (right)

\subsubsection{Planning in the Latent Space}
Once we have an approximate linear latent dynamics model, all that remains is to encode the planning problem as a trajectory optimization problem in the latent space. 

More precisely, given a trajectory planning problem, our aim is to find an approximate trajectory planning problem in the latent space:
\[
\begin{aligned}
    \text{min } &J(\bm{x}, \bm{u}) \\
    \text{s.t. } &\bm{x}(0) = \bm{x}_{\text{initial}} \\
    &\bm{x}_{t+1} = f(\bm{x}_t, \bm{u}_t) \\
    & H\bm{x} \leq \bm{g} \\
    &\bm{x}(T) = \bm{x}_{\text{goal}}
\end{aligned} \longrightarrow
\begin{aligned}
    \text{min } &J(\bm{z}_x, \bm{z}_u) \\
    \text{s.t. } &\bm{z}(0) = \bm{\varphi}_x (\bm{x}_{\text{initial}}) \\
    &\bm{z}_{t+1} = K\bm{z}_t \\
    & \mathcal{H}\bm{z} \leq \mathpzc{g} \\
    &\bm{z}(T) = \bm{\varphi}_x(\bm{x}_{\text{goal}})
\end{aligned}
\]
We assume that the state constraints given by $H\bm{x} \leq \bm{g}$ describe convex polytopal regions. Safety constraints computed by the trajectory validator can be represented in this way as well. Using our accurate VAE dynamics model, we can easily encode the initial state and goal state into the latent space. For this encoding, and for the rest of the section, we will use the mean of the latent distribution as the latent vector (instead of sampling) for repeatability.

Next, we encode each convex polytopal constraint into the latent space by encoding each vertex. Note that when we encode the polytope vertices into the latent space, there is not guarantee that the resulting latent polytope will be convex. This is because the latent space is only \textit{approximately} linear. To address this challenge, we approximate the polytope with a hyperrectangle computed from the extremes of all the latent vertices in each dimension. Because the trajectory validator will assess the safety of the trajectory in the real space using the exact constraints and obstacles, the trajectory planner can use approximations in candidate generation.

Finally, we use trajectory optimization to generate an optimal control trajectory in latent space $\bm{z}_u^*$. We then decode this trajectory to get the candidate trajectory $\bm{u}^*$ in real space. This candidate trajectory is passed to the trajectory validator for risk assessment.
\section{Results}
In this section, we compare our planning algorithm to the state of the art in risk-bounded trajectory planning and demonstrate planning with learned models using a case study on real-world data of an under-actuated autonomous system. Our risk-bounded planning algorithm is implemented in Python and all experiments were run on a 4.5GHz processor using GUROBI 8.0.1 as the optimizer for the trajectory planner.

\subsection{Planning Algorithm Performance}

We compare our algorithm to the state of the art in risk-bounded trajectory planning algorithm for nonlinear stochastic agents \cite{weiqiao2022}. For the sake of comparison, we run our planning algorithm in real-space, using close-form stochastic dynamics to sample trajectories for the trajectory validator. 

Consider an autonomous nonlinear agent with stochastic dynamics given by

\begin{align} 
\begin{split}
x_{t+1}&=x_{t}+\Delta t (v_{t}+\omega_{v_t} )cos\left(\theta_{t}+ \omega_{\theta_t}\right) \\
y_{t+1}&=y_{t}+\Delta t (v_{t}+\omega_{v_t} )sin\left(\theta_{t} + \omega_{\theta_t}\right)\\
\end{split}
\label{eqn:dubin_dyn}
\end{align}

where $(x, y)$ are the state variables representing position, $(v, \theta)$ are the control variables representing velocity and steering angle, and the time step $\delta t = 0.1$. The noise is modeled using random variables $\omega_{v_t}$ and $\omega_{\theta_t}$ with a uniform distribution over [-0.1, 0.1]. The initial state is given my a uniform distribution over [-0.1, 0.1] x [-0.1, 0.1]. The objective function to be minimized is $\sum_{t=0}^{T} v_t^2 + \theta_t^2$, the risk bound $\delta = 0.1$, and the goal region is centered on [1, 0.5]. Each planning algorithm was run for 100 trials, varying the initial condition.

\begin{figure}
    \centering
    \includegraphics[width=0.45\textwidth]{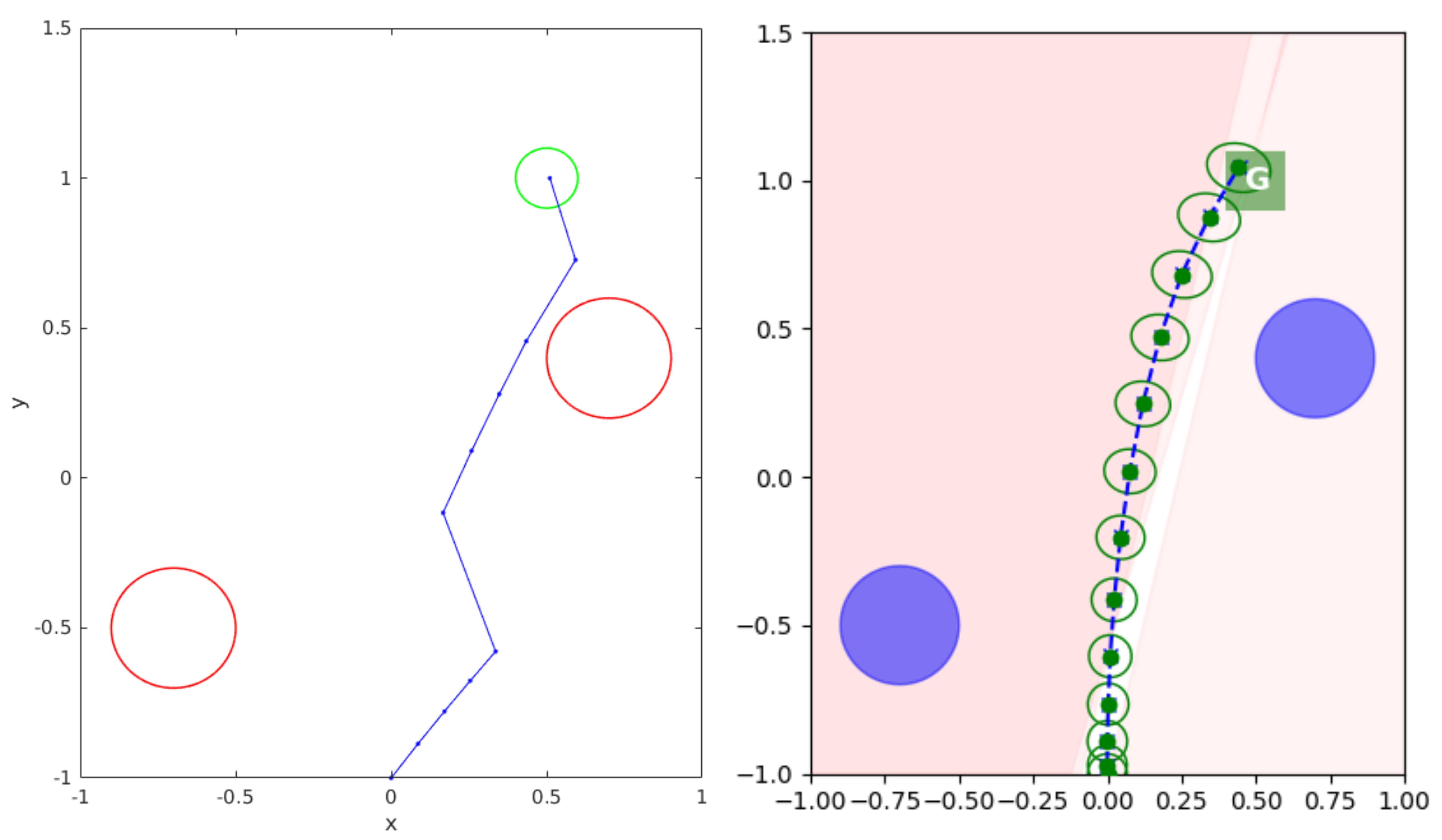}
    \caption{Planner comparison. (left) Example trajectory generated by \cite{weiqiao2022}'s approach. (right) Example trajectory by our approach. }
    \label{fig:compare}
\end{figure}

\figref{fig:compare} shows the solution trajectory generated by \cite{weiqiao2022} (left) and the solution trajectory generated by our method (right). On average, our algorithm was able to generate a safe trajectory in around 11 seconds (5 iterations) while it took \cite{weiqiao2022} over 120 seconds. Our solution had an average total risk of 0.032 and an average objective value of 85.28 compared to \cite{weiqiao2022}'s solution with an average total risk of 0.023 and an average objective value of 74.16. Our solution shows an order of magnitude improvement in planning speed with only a $10\%$ loss in optimality and safety.

\subsection{Case Study: Planning from Drone Data}
To demonstrate that our algorithm can generate safe plans for agents with learned behaviors, we performed a case study using an open source drone dataset. The Blackbird drone dataset is a large-scale suite of sensor and ground truth data for a custom-built autonomous quadrotor equipped with an IMU and rotor tachometers \cite{blackbird}. The dataset contains over 10 hours of flying at various maximum speeds through 18 different trajectories. There is no closed-form dynamics model for the quadrotor provided with the dataset.

The Blackbird quadrotor has 6 state variables $(\begin{smallmatrix}
x & y & z & roll & pitch & yaw
\end{smallmatrix})$ and 4 control variables (rotor speed for each rotor). For this dataset, the learned VAE had a latent state dimension of 9 and a latent control dimension of 7 (determined through tuning). The higher dimensional of the latent space allows the learned model to account for the nonlinearities in the dynamics using additional features. Each the encoder and decoder for each VAE had four layers each with 500 nodes. 

\begin{figure}
    \centering
    \includegraphics[width=0.45\textwidth]{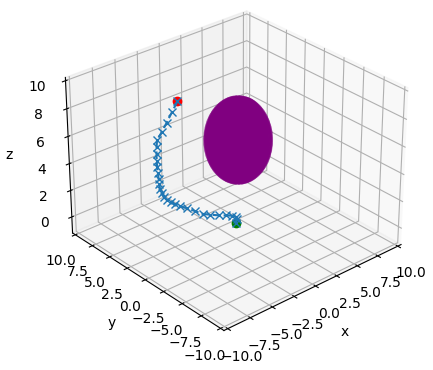}
    \caption{Trajectory planned for the Blackbird quadrotor using learned dynamics. The green and red circles are the initial state and goal state. The purple ellipsoid is an obstacle.}
    \label{fig:blackbird}
\end{figure}

As shown in \figref{fig:blackbird}, our algorithm was able to generate a safe trajectory plan for the quadrotor without prior knowledge of the dynamics. The final trajectory respects the control limits of the quadrotor, avoids the ellipsoidal obstacles, and takes the agent from the initial state to the goal.

\section{Conclusion}

In this work, we presented LaPlaSS: Latent space Planning for Stochastic Systems, an efficient risk-bounded planning algorithm for autonomous mobile agents with learned dynamics. LaPlaSS takes a "generate-and-test" approach in which a trajectory planner generates a candidate trajectory using approximate linear deterministic dynamics. Then, a trajectory validator assesses the risk of the candidate trajectory by computing a probabilistic flow tube using trajectories sampled from an accurate dynamics model. The trajectory validator returns a safety constraint to the trajectory planner if the candidate does not meet the provided risk bound. To acquire an accurate dynamics model, we use a variational autoencoder model to capture the uncertainties in dynamics. We then learn a latent linear dynamics model and encode the planning problem into the latent space to generate the candidate trajectory. Our results show that our planning algorithm is an order of magnitude more efficient than the state of the art and demonstrate that we can generate a safe trajectory for a real-world agent with learned dynamics.

\bibliographystyle{IEEEtran}
\bibliography{bare_jrnl}
%

%








\end{document}